\def\year{2020}\relax
\documentclass[letterpaper]{article} 
\usepackage{aaai20}  
\usepackage{times}  
\usepackage{helvet} 
\usepackage{courier}  
\usepackage[hyphens]{url}  
\usepackage{graphicx} 
\usepackage{comment}
\usepackage{graphicx}  
\usepackage{multirow}

\title{Emergency Department Optimization and Load Prediction in Hospitals}
\author{\Large \textbf{Karthik K. Padthe,\textsuperscript{1}
Vikas Kumar,\textsuperscript{1}
Carly M. Eckert MD MPH,\textsuperscript{1,2}
Nicholas M. Mark MD,\textsuperscript{1,3}}\\
\\ \Large \textbf{Anam Zahid,\textsuperscript{1}
Muhammad Aurangzeb Ahmad,\textsuperscript{1,4}
Ankur Teredesai,\textsuperscript{1,5} }\\
\textsuperscript{1}{KenSci Inc, Seattle, WA}\\
\textsuperscript{2}{Department of Epidemiology, University of Washington}\\
\textsuperscript{3}{Swedish Medical Center, Seattle, WA}\\
\textsuperscript{4}{Department of Computer Science, University of Washington - Bothell}\\
\textsuperscript{5}{Department of Computer Science, University of Washington - Tacoma}
\\ 
\\ 
\{karthik, vikas, carly, drnick, anam, muhammad, ankur\}@kensci.com \\ 
}

\begin{document}

\maketitle

\begin{abstract}
Over the past several years, across the globe, there has been an increase in people seeking care in emergency departments (EDs).  ED resources, including nurse staffing, are strained by such increases in patient volume. Accurate forecasting of incoming patient volume in emergency departments (ED) is crucial for efficient utilization and allocation of ED resources. Working with a suburban ED in the Pacific Northwest, we developed a tool powered by machine learning models, to forecast ED arrivals and ED patient volume to assist end-users, such as ED nurses, in resource allocation. 
In this paper, we discuss the results from our predictive models, the challenges, and the learnings from users' experiences with the tool in active clinical deployment in a real world setting. 
\end{abstract}

\section{Introduction}
\begin{figure} [t]
 \includegraphics[height=4cm, width=8.5cm]{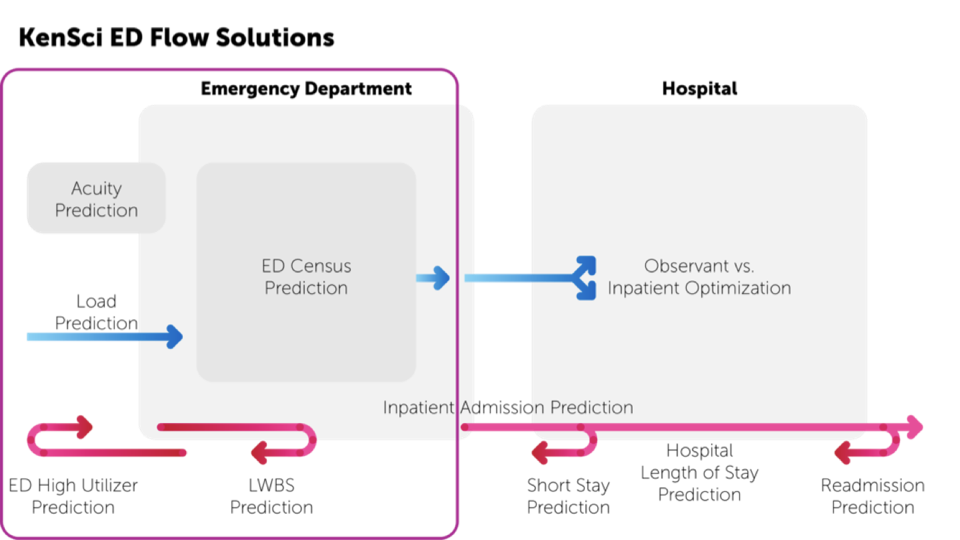}
  \caption{Overview of set of prediction models that can help optimize Emergency Department efficiency.}
  \label{fig:edflow}
\end{figure}

Emergency departments (EDs) are a critical component of  the healthcare infrastructure and ED crowding is a global problem. In 2016 there were over 140 million ED visits in the US ~\cite{CDC2015emergency}. The number of ED patients is growing and, according to US data, this increase has outpaced population growth for the last 20 years~\cite{weiss2006overview}. As a result, EDs are increasingly crowded  ~\cite{mccarthy2008challenge} and ED overcrowding has been linked to decreased quality of care~\cite{schull2003emergency}~\cite{hwang2006effect}, increased costs~\cite{bayley2005financial}, and increased patient dissatisfaction~\cite{jenkins1998violence}.  Using machine learning models to predict ED load could ameliorate the adverse effects of crowding, and multiple strategies have been proposed, including forecasting future crowding ~\cite{Hoot_LeBlanc_Jones_Levin_Zhou_Gadd_Aronsky_2009}, predicting the likelihood of inpatient admission ~\cite{Peck_Benneyan_Nightingale_Gaehde_2012}, and predicting the likelihood that a patient will leave the ED without being seen ~\cite{pham2009national}. These solutions use a variety of administrative and patient level data to attempt to mitigate common ED bottlenecks, bottlenecks that uncorrected may lead to delays, inefficiencies, and even deaths ~\cite{carter2014relationship}. Multiple factors influence ED crowding including the number of new patients coming to the ED (arrivals), how severely sick or injured patients are (acuity), and the total number of patients in the ED (census). Each of these factors have both stochastic and deterministic components ~\cite{jones2009multivariate}~\cite{jones2008forecasting} and are influenced by both exogenous (e.g., vehicle crashes) and endogenous factors (e.g., hospital processes). In order to optimize ED flow, it is therefore necessary to integrate multiple predictions as shown in Figure~\ref{fig:edflow}.

If ED load could be accurately predicted, staffing could be adjusted to optimize patient care. The ability to predict the number of patients seeking ED care on a given day is essential to optimizing nurse staffing ~\cite{batal2001predicting}. Currently, ED nurse staffing is assigned using heuristics and anecdotes such as higher census on Mondays, on days following federal holidays, and with other factors such as changes in weather, traffic, and local sporting events. Inaccurate prediction can lead to inappropriate nurse to patient ratios which can lead to dangerous under-staffing, poor clinical outcomes, nursing dissatisfaction, and burnout \cite{aiken2002hospital}. Matching staffing levels to the variation in daily patient demand can improve the quality of care and lead to cost savings. 

\section{Related Work}

In this paper, we present our work with a busy suburban ED in the Pacific Northwest that services a rapidly growing metropolitan area. We describe the development of novel models to predict ED arrivals and census, the design of an easily consumable dashboard integrated into the clinical workflow, and deployment of the dashboard using a live data feed. The current work also addresses a gap in the literature where there is a dearth of published work related to ED optimization in a real world setting and in production.


The availability of accessible data and computational resources has enabled the application of machine learning (ML) to healthcare at an unprecedented scale ~\cite{Krumholz_2014}. While several research groups have developed ML predictions on retrospective and static ED data, operationalized ML solutions in the ED are rare.
Chase et al. developed a novel indicator of a busy ED: a care utilization ratio~\cite{chase2012predicting}. The authors report that the prediction of this ratio, which incorporates new ED arrivals, number of patients triaged, and physician capacity, provides a robust indicator of ED crowding. McCarthy et al. utilized a Poisson regression model to predict demand for ED services~\cite{mccarthy2008challenge}. They determined that after accounting for temporal, weather, and patient-related factors (hour of day is most important), ED arrivals during one hour had little to no association with the number of ED arrivals the following hour.  Jones et al. ~\cite{jones2008forecasting} explored seasonal autoregressive integrated moving average (SARIMA), time series regression, exponential smoothing, and artificial neural network models to forecast daily patient volumes and also identified seasonal and weekly patterns in ED utilization.

\section{ED Predictions}
The goal of our work was to optimize ED operations by accurately predicting ED arrivals and ED patient census to facilitate staffing optimization to better manage the influxes and patterns of ED patients to provide safe and timely care. Here we describe our approach to building the prediction models and we describe the metrics we used to evaluate the model accuracy.

\subsection{Problem Description}
\label{problemSetup}
There are two distinct yet related ED load optimization problems that we address in this work, as described below:

\subsubsection{ED Census}
ED census is defined as the total number of patients in the ED at a specified time. ED census includes patients in the waiting room, in triage, those receiving care, and those awaiting ED disposition: hospital admission, discharge, or transfer. ED census is a "snapshot" of ED utilization and includes elements related to ED arrivals as well as ED throughput. Predicting ED census can serve to inform both short-term (minutes to hours) operations, such as reassigning staff or diverting ambulance arrivals and longer-term (hours or longer) administrative decisions, such as calling in additional staff or sending staff members home early. We formulated this problem as a prediction of ED census at $t+2 ~hours$, $t+4~hours$, and $t+8~hours$, where $t$ is the prediction time. In production, these predictions are made every $15~minutes$, resulting in near real-time predictions. For instance, at 3:15 PM ($t$), we predict census for 5:15 PM ($t+2~hours$), 7:15 PM ($t+4~hours$), and 11:15 PM ($t+8~hours$). Then at 3:30 PM ($t$), we predict 5:30 PM ($t+2~hours$), 7:30 PM ($t+4~hours$), and 11:30 PM ($t+8~hours$).

\begin{table}[t]
\centering
\begin{small}
\caption{ED Arrivals and Census Features. The \textit{prior census} and \textit{slope Census} are used only in census prediction model and \textit{prior arrival} and \textit{slope arrival} only in arrivals model.}
\label{tbl:features}
\begin{tabular}{@{}cp{5cm}@{}}
\hline
\textbf{Feature} & \textbf{Description}    \\
\hline
\textbf{Prior Census/Arrival}  & 4 features; census/arrival at 4 time events (at 15 min intervals) prior to prediction time, i.e. 15 min, 30 min, 45 min, 60 min.\\

\textbf{Month of year} & January - December (12 features)                                                                                                  \\

\textbf{Hour of day}  & Hour of the day (24 features)\\

\textbf{Day of Week} & Day of the week (7 features)\\

\textbf{Quarter of Year} & Season: Q1 Winter, Q2 Spring, Q3 Summer, Q4 Autumn \\
\textbf{Weekend Flag} & Flag if prediction on Saturday or Sunday\\
\textbf{Evening Flag} & Flag if prediction time between 20:00 and 08:00\\
\textbf{Slope census/arrivals}  & Slope of change from prior census or arrival\\

\hline
\end{tabular}
\end{small}
\end{table}

\subsubsection{ED Arrivals and Acuity}
ED arrivals reflect the number of individual patients who are arriving at the ED over a period of time.  Arrivals can be described by the acuity level of the individual patient, an indicator of illness or injury severity assessed by nursing staff at the time of patient triage~\cite{gilboy2012emergency}.  Predictions of patient volume by acuity level can further inform staffing needs - higher acuity patients tend to have greater intensity of staff and resource needs.  Similar to the \textit{Census} prediction, we framed the \textit{Arrivals} prediction by acuity for 2, 4, and 8 hour forecasting. To accommodate different patterns in the acuity of patients, we built models for each individual acuity level.

\section{Methods}
For both \textit{Census} and \textit{Arrivals} we include temporal features such as hour of day, day of week, month of year, and quarter of year. To include the unique variations in census and arrival patterns in the evening compared to the morning as well as weekend versus weekday patterns, we included corresponding binary variables.

While ED census or ED arrival may be independent from one hour to the next, we use the current ED census \textit{trend} to inform future ED census. To include signals for the current census trends in ED in our predictive models, we determine the slope from the census values in the previous 1 hour for every 15 minute intervals. In addition, we weighted values from these 15 minute intervals to that more recent values had higher weights. The census at $t-15~minutes$, $t-30~minutes$, $t-45~minutes$, and  $t-60~minutes$ is weighted with 2, 0.5, 0.25, and 0.05 respectively. The weights were chosen empirically based on the performance metrics of the model. Similar to \textit{Census}, the arrivals for the \textit{Arrival} prediction are weighted in the same way. The final set of features is shown in Table~\ref{tbl:features}.

\begin{table}[t]
\centering
\begin{small}
\caption{Distribution of ED Encounters by Acuity}
\label{tbl:data}
\begin{tabular}{@{}lcc@{}}
\hline
\textbf{Acuity} & \textbf{ESI} & \textbf{Number of Encounters} \\
\hline
\textbf{Emergent} & 1 & 1,435 \\
     &  2  & 46,436 \\
\textbf{Urgent} & 3 & 116,808 \\
\textbf{Non-Urgent} & 4 & 33,023 \\
     &  5 & 2,315 \\
  \hline
  \textbf{Total} & & \textbf{199,957}\\
\hline
\end{tabular}
\end{small}
\end{table}

\vspace{-2pt}
\subsection{Dataset Description}
The data for the experiments came from a suburban level three trauma center at a hospital in the Pacific Northwest with $>60,000$ annual ED visits.  The ED comprises multiple treatment spaces including 40 acute treatment rooms and 4 trauma rooms for the resuscitation of critically ill patients.  Individuals are registered at the time of entry to the ED and all registered ED patients were included in this analysis. ED encounters occurring between January 2014 through January 2018 were included in the experiments.  The dataset included electronic health record (EHR) data elements such as time, date, location, chief complaint, acuity score, vital signs, and others. This included $205,929$ ED encounters, of which $199,957$ encounters documented patient acuity. ESI is a categorical variable representing patient acuity (based on vital signs and symptoms) where ESI 1 connotes highest urgency and ESI 5 the lowest urgency \cite{gilboy2012emergency}. We grouped these into three categories reflecting emergent (ESI 1 or 2), urgent (ESI 3), and non urgent (ESI 4 or 5). The distribution of of the encounters split by ESI groups is shown in Table~\ref{tbl:data}.

\subsection{Models}
Multiple regression models were evaluated for both \textit{Census} and \textit{Arrivals} predictions. We choose to use a Generalized Linear Model with Poisson Regression (GLM) for its simplicity and capability to model count data~\cite{gardner1995regression}. We included regularization variants of GLM that include Lasso, Ridge, and Elastic Net for validation. We also included linear Gradient Boosting Machine (GBM) due to its robustness to missing data and predictive power~\cite{friedman2001greedy}. We used the average arrivals and census values at that same time point from the prior two years as our baseline. We used scikit-learn package available in Python 3.6 to implement all models. 

\subsection{Evaluation metrics}
We evaluate the performance of our models using root mean squared error (RMSE) and  mean absolute error (MAE) \cite{verbiest2014evaluation} which are suitable metrics for regression. However, the real utility of ED load prediction is in staffing optimization. Most common mid-size US ED departments have an ED patient to nurse ratio of 4:1.  Based on this, we devised an additional metric: we determined the percentage of times the model prediction is within a threshold of $\pm 4$ (\textit{Absolute Error} $<=4$). Furthermore, we also calculate the percentage of times that the model is accurate to within $70\%$ of the actual value (\textit{Accuracy}$>$70\%). These additional metrics frame the models performances in terms of their effects on user workflows and provide a simple understanding of the model performance under the system constraints while ensuring interpretability to end users.

\begin{figure} [t]
  \includegraphics[width=\columnwidth]{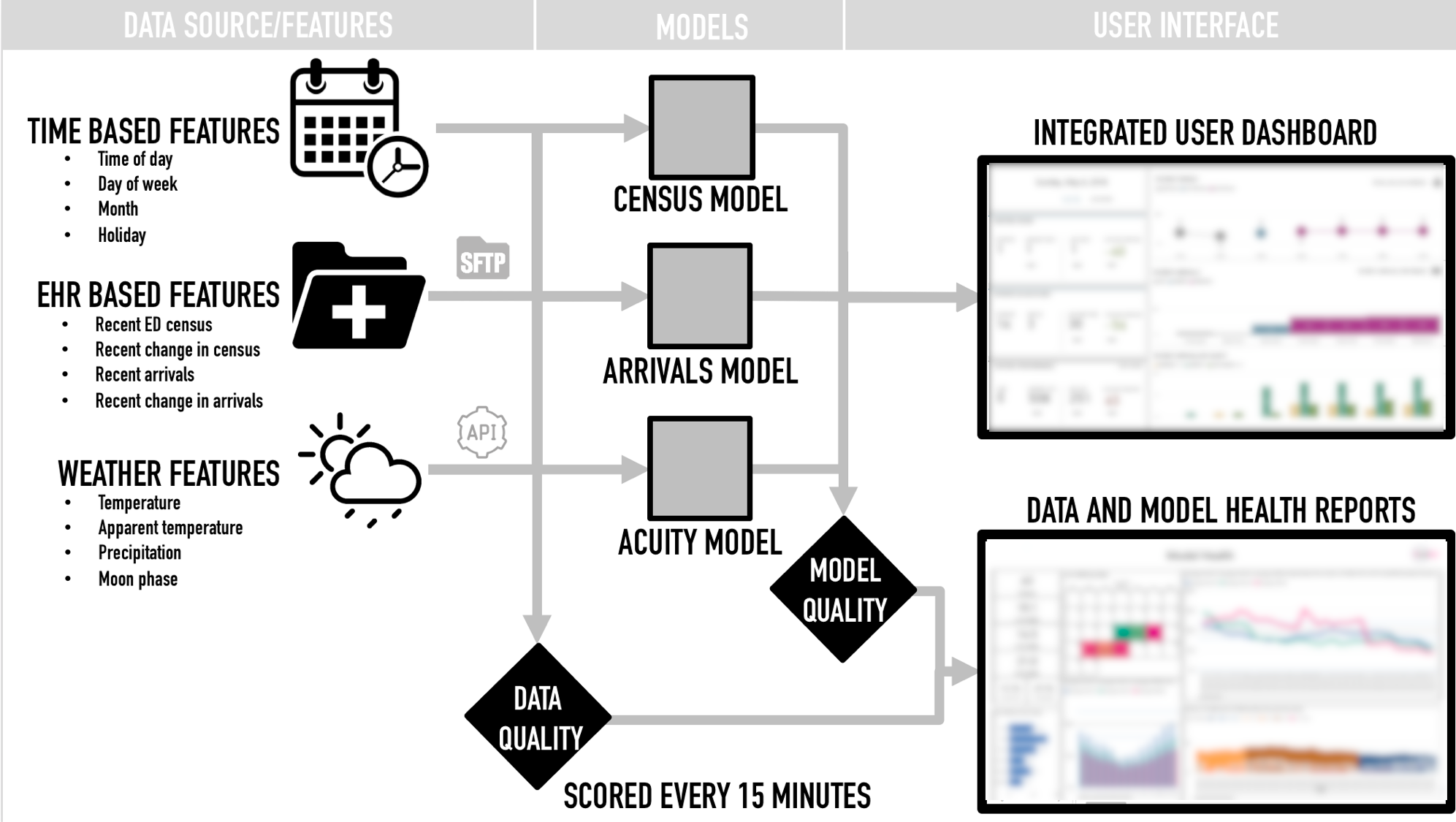}
  \caption{Schematic showing the data sources, models, and resulting User and Model Health Dashboards. The actual dashboard image is hidden due to privacy and data compliance.}
  \label{fig:boat1}
\end{figure}

\begin{figure} [t]
 \includegraphics[height=4cm, width=8.5cm]{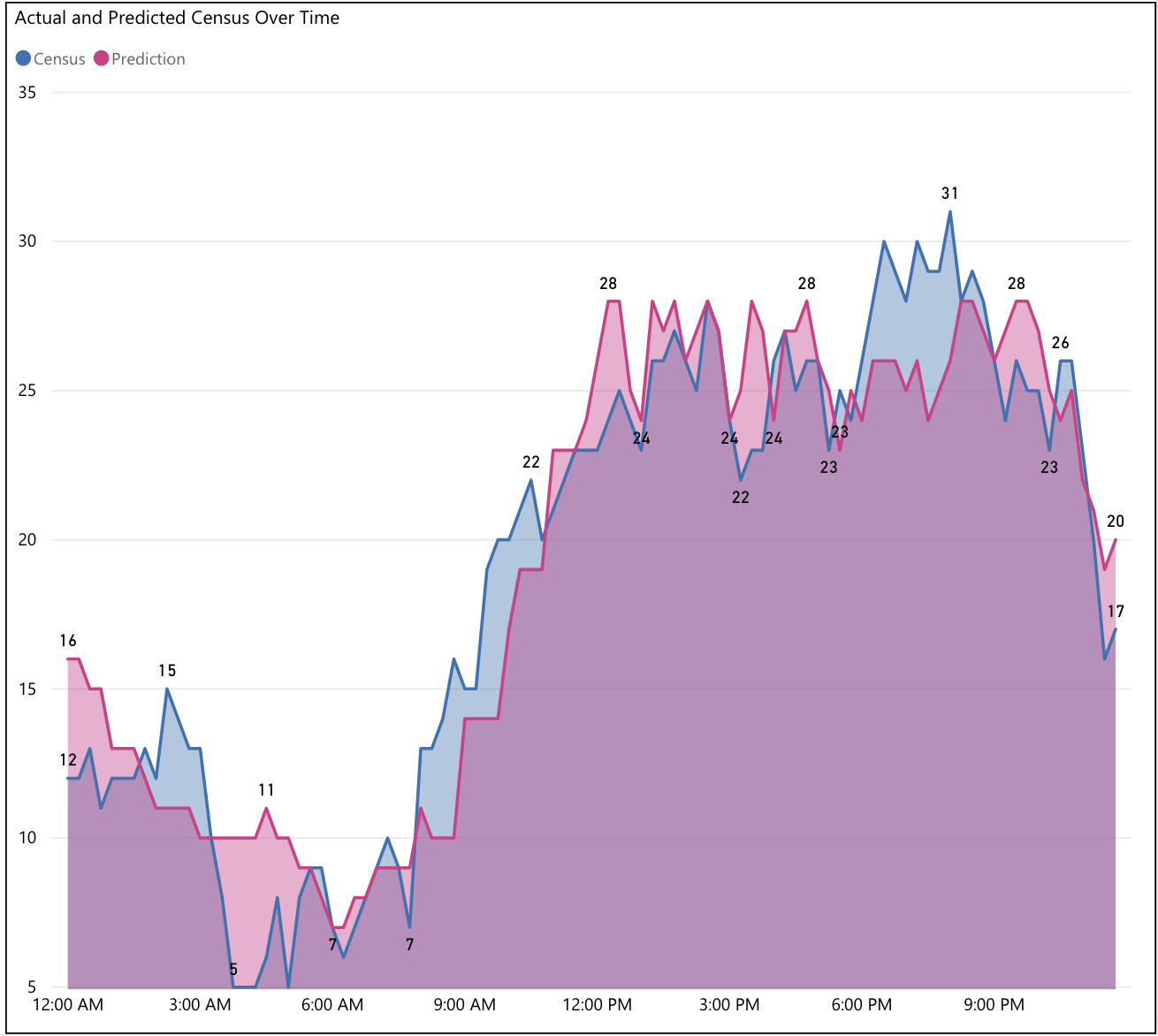}
  \caption{Monitoring model performance in deployment - Example of predicted vs actual census for the 2 hour census prediction model over the course of one day}
  \label{fig:boat2}
\end{figure}

Furthermore, combining these models with a model management process to detect changes in model performance or shifts in underlying patient distributions, prevails as novel work. Model management is an iterative process that includes monitoring and evaluating model performance to detect subtle (or unsubtle) changes in the underlying distribution of the data, permitting investigation and, if necessary, model re-training. We have  implemented a workflow for automatic model monitoring; the overview of this is represented in Figure~\ref{fig:boat1}. As part of this workflow we created a user friendly dashboard to track the model performance and distributions, an example visual can be seen in Figure~\ref{fig:boat2}.

\subsection{Results}
Data from January 2014 to October 2017 was used to train the models and  data from November 2017 to January 2018 was used to test the models. The performance metrics of the census models for 2 hour prediction are shown in Table~\ref{tbl:results}. The Gradient Boosting Method (GBM) performed the better among the set for all metrics which we believe is due to its robustness to the sparsity in the data. The 4 and 8 hours GBM census model MAEs are $4.0739$ and $4.2960$ respectively. The metric (Accuracy $>$ 70\%) shows that GBM is accurate 81.52\% of times for a prediction within 70\% of actual census. And, the GBM is accurate 72.90\% time for a prediction within a value of $\pm4$ of actual census.

\begin{table*}[]
\centering
\caption{Results of 2, 4 and 8 hour Arrival prediction for GLM variants, GBM, and Baseline model for Emergent, Urgent and Non-urgent patients}
\label{tbl:results_arrival}
\begin{tabular}{lllllcc}
\hline
Acuity & Time window & Model & RMSE & MAE & \begin{tabular}[c]{@{}l@{}}Absolute  Error\textless{}4\end{tabular} & \begin{tabular}[c]{@{}l@{}}Accuracy \textgreater{}70\%\end{tabular} \\ \hline
\multicolumn{1}{l|}{\multirow{17}{*}{Emergent}} & \multicolumn{1}{l|}{\multirow{6}{*}{2 hour}} &\textbf{GLM} & 1.9747 & 1.4267 & 96.33 & 32.80\\
\multicolumn{1}{l|}{} & \multicolumn{1}{l|}{} & {GLM-Lasso} & 2.1492 & 1.5400 & 94.96 & 31.96\\
\multicolumn{1}{l|}{} & \multicolumn{1}{l|}{} & {GLM-Ridge} & 2.0039 & 1.4451 & 96.05 & 32.45\\
\multicolumn{1}{l|}{} & \multicolumn{1}{l|}{} & {GLM-Elastic Net} & 2.1494 & 1.5396 & 95.08 & 31.91\\
\multicolumn{1}{l|}{} & \multicolumn{1}{l|}{} & \textbf{GBM} & 1.9768 & 1.4283 & 96.38 & 32.97\\
\multicolumn{1}{l|}{} & \multicolumn{1}{l|}{} & \textbf{Baseline} &  2.0278 & 1.5174 & 96.59 & 21.63 \\ \cline{2-7} 
\multicolumn{1}{l|}{} & \multicolumn{1}{l|}{\multirow{6}{*}{4 hour}} & \textbf{GLM} & 1.9749 & 1.4272 & NA & NA\\
\multicolumn{1}{l|}{} & \multicolumn{1}{l|}{} & {GLM-Lasso} & 2.1913 & 1.5642 & NA & NA\\
\multicolumn{1}{l|}{} & \multicolumn{1}{l|}{} & {GLM-Ridge} & 2.0318 & 1.4615 & NA & NA\\
\multicolumn{1}{l|}{} & \multicolumn{1}{l|}{} & {GLM-Elastic Net} & 2.1700 & 1.5467 & NA & NA\\
\multicolumn{1}{l|}{} & \multicolumn{1}{l|}{} & \textbf{GBM} & 1.9786 & 1.4276 & NA & NA\\
\cline{2-7} 
\multicolumn{1}{l|}{} & \multicolumn{1}{l|}{\multirow{5}{*}{8 hour}} & \textbf{GLM} & 2.9998 & 2.1928 & NA & NA\\
\multicolumn{1}{l|}{} & \multicolumn{1}{l|}{} & {GLM-Lasso} & 3.5831 & 2.6451 & NA & NA\\
\multicolumn{1}{l|}{} & \multicolumn{1}{l|}{} & {GLM-Ridge} & 3.1154 & 2.2680 & NA & NA\\
\multicolumn{1}{l|}{} & \multicolumn{1}{l|}{} & {GLM-Elastic Net} & 3.4307 & 2.5060 & NA & NA\\
\multicolumn{1}{l|}{} & \multicolumn{1}{l|}{} & \textbf{GBM} & 3.0391 & 2.2217 & NA & NA\\
\hline
\multicolumn{1}{l|}{\multirow{17}{*}{Urgent}} & \multicolumn{1}{l|}{\multirow{6}{*}{2 hour}} & \textbf{GLM} & 1.5022 & 1.1088 & NA & NA \\
\multicolumn{1}{l|}{} & \multicolumn{1}{l|}{} & {GLM-Lasso} & 1.5837 & 1.1576 & NA & NA \\
\multicolumn{1}{l|}{} & \multicolumn{1}{l|}{} & {GLM-Ridge} & 1.5116 & 1.1168 & NA & NA \\
\multicolumn{1}{l|}{} & \multicolumn{1}{l|}{} & {GLM-Elastic Net} & 1.5630 & 1.1451 & NA & NA \\
\multicolumn{1}{l|}{} & \multicolumn{1}{l|}{} & {GBM} & 2.4042 & 1.6891 & NA & NA \\
\cline{2-7} 
\multicolumn{1}{l|}{} & \multicolumn{1}{l|}{\multirow{6}{*}{4 hour}} & \textbf{GLM} & 1.5010 & 1.1082 & NA & NA \\
\multicolumn{1}{l|}{} & \multicolumn{1}{l|}{} & {GLM-Lasso} & 1.5883 & 1.1511 & NA & NA \\
\multicolumn{1}{l|}{} & \multicolumn{1}{l|}{} & {GLM-Ridge} & 1.5065 & 1.1102 & NA & NA\\
\multicolumn{1}{l|}{} & \multicolumn{1}{l|}{} & {GLM-Elastic Net} & 1.5514 & 1.1311 & NA & NA\\
\multicolumn{1}{l|}{} & \multicolumn{1}{l|}{} & {GBM} & 2.4066 & 1.6914 & NA & NA \\
\cline{2-7} 
\multicolumn{1}{l|}{} & \multicolumn{1}{l|}{\multirow{5}{*}{8 hour}} & \textbf{GLM} & 2.1792 & 1.6305 & NA & NA \\
\multicolumn{1}{l|}{} & \multicolumn{1}{l|}{} & {GLM-Lasso} & 2.4764 & 1.8841 & NA & NA \\
\multicolumn{1}{l|}{} & \multicolumn{1}{l|}{} & {GLM-Ridge} & 2.2214 & 1.6917 & NA & NA \\
\multicolumn{1}{l|}{} & \multicolumn{1}{l|}{} & {GLM-Elastic Net} & 2.3984 & 1.8109 & NA & NA \\
\multicolumn{1}{l|}{} & \multicolumn{1}{l|}{} & {GBM} & 3.9960 & 2.8792 & NA & NA \\ \hline
\multicolumn{1}{l|}{\multirow{17}{*}{Non-Urgent}} & \multicolumn{1}{l|}{\multirow{6}{*}{2 hour}} & \textbf{GLM} & 2.5903 & 2.0017 & NA & NA \\
\multicolumn{1}{l|}{} & \multicolumn{1}{l|}{} & {GLM-Lasso} & 2.6859 & 2.0874 & NA & NA \\
\multicolumn{1}{l|}{} & \multicolumn{1}{l|}{} & {GLM-Ridge} & 2.5911 & 1.9996 & NA & NA \\
\multicolumn{1}{l|}{} & \multicolumn{1}{l|}{} & {GLM-Elastic Net} & 2.6572 & 2.0412 & NA & NA \\
\multicolumn{1}{l|}{} & \multicolumn{1}{l|}{} & {GBM} & 4.0945 & 3.5052 & NA & NA \\ \cline{2-7} 
\multicolumn{1}{l|}{} & \multicolumn{1}{l|}{\multirow{6}{*}{4 hour}} & \textbf{GLM} & 2.5987 & 2.0069 & NA & NA \\
\multicolumn{1}{l|}{} & \multicolumn{1}{l|}{} & {GLM-Lasso} & 2.7321 & 2.0956 & NA & NA \\
\multicolumn{1}{l|}{} & \multicolumn{1}{l|}{} & {GLM-Ridge} & 2.5989 & 2.0038 & NA & NA \\
\multicolumn{1}{l|}{} & \multicolumn{1}{l|}{} & {GLM-Elastic Net} & 2.7022 & 2.0918 & NA & NA \\
\multicolumn{1}{l|}{} & \multicolumn{1}{l|}{} & {GBM} & 4.1064 & 3.5214 & NA & NA \\
\cline{2-7} 
\multicolumn{1}{l|}{} & \multicolumn{1}{l|}{\multirow{5}{*}{8 hour}} & \textbf{GLM} & 3.7944 & 2.9291 & NA & NA \\
\multicolumn{1}{l|}{} & \multicolumn{1}{l|}{} & {GLM-Lasso} & 4.2303 & 3.2661 & NA & NA \\
\multicolumn{1}{l|}{} & \multicolumn{1}{l|}{} & {GLM-Ridge} & 3.8033 & 2.9463 & NA & NA \\
\multicolumn{1}{l|}{} & \multicolumn{1}{l|}{} & {GLM-Elastic Net} & 4.3284 & 3.3386 & NA & NA \\
\multicolumn{1}{l|}{} & \multicolumn{1}{l|}{} & {GBM} & 7.6992 & 6.8340 & NA & NA \\ \hline
\end{tabular}
\end{table*}

For arrival models, we built 9 models, one for each acuity level and for each 2, 4 and 8 hours prediction. We observed that the gradient boosting model performed better than other models and the baseline for Emergent acuity encounters, where as for Urgent, Non-urgent acuity GLM models performed better. The results are shown in Table~\ref{tbl:results_arrival}. The absolute error and accuracy were only available for a subset of models. We observe that the MAE and RMSE for all models across different levels of acuity is similar if we consider 2 hour and 4 hour windows. However, the performance goes down if we consider 8 hour windows. This is not unexpected since trends can greatly vary across longer time spans e.g., compare ED trends at 2 am vs. 10 am.

\begin{table*}[]
\centering
\caption{Results of 2, 4 and 8 hour Census prediction for GLM variants, GBM, and Baseline model. The baseline is average of census at that hour in previous 2 years.}
\label{tbl:results}
\begin{tabular}{llcccc}
\hline
Time window                                  & Model & RMSE & MAE & \begin{tabular}[c]{@{}l@{}}Absolute Error\textless{}4\end{tabular} & \begin{tabular}[c]{@{}l@{}}Accuracy \textgreater{}70\%\end{tabular} \\ \hline
\multicolumn{1}{l|}{\multirow{6}{*}{2 hour}} & \textbf{GLM} & 4.3343 & 3.3812 &  71.80 & 80.42\\
\multicolumn{1}{l|}{} & {GLM-Lasso} & 4.6816 & 3.6642 & 68.41 & 77.83\\
\multicolumn{1}{l|}{} & {GLM-Ridge} & 4.5305 & 3.4975 & 70.79 & 72.80\\
\multicolumn{1}{l|}{} & {GLM-Elastic Net} & 4.6550 & 3.6395 & 69.45 & 78.47\\
\multicolumn{1}{l|}{} & \textbf{GBM} & 4.2013 & 3.2790 & 72.90 & 81.52\\
\multicolumn{1}{l|}{} &  \textbf{Baseline} & 6.9026 & 5.3926 & 51.19 & 60.19 \\
\hline
\multicolumn{1}{l|}{\multirow{5}{*}{4 hour}} & \textbf{GLM(4 hour)} & 5.1491 & 4.0173 &  64.66 & 74.95\\
\multicolumn{1}{l|}{} & {GLM-Lasso(4 hour)} & 6.0410 & 4.6643 & 56.92 & 68.78\\
\multicolumn{1}{l|}{} & {GLM-Ridge(4 hour)} & 5.2855 & 4.1111 & 63.85 & 74.44\\
\multicolumn{1}{l|}{} & {GLM-Elastic Net(4 hour)} & 6.1962 & 4.7468 & 57.32 & 67.59\\
\multicolumn{1}{l|}{} & \textbf{GBM(4 hour)} & 5.1784 & 4.0241 & 64.35 & 75.05\\
\hline
\multicolumn{1}{l|}{\multirow{5}{*}{8 hour}} & \textbf{GLM(8 hour)} & 5.5026 & 4.2960 &  61.42 & 72.51\\
\multicolumn{1}{l|}{} & {GLM-Lasso(8 hour)} & 6.1726 & 4.8158 & 55.62 & 67.26\\
\multicolumn{1}{l|}{} & {GLM-Ridge(8 hour)} & 5.5829 & 4.3632 & 60.68 & 72.25\\
\multicolumn{1}{l|}{} & {GLM-Elastic Net(8 hour)} & 6.6123 & 5.1085 & 53.79 & 64.54\\
\multicolumn{1}{l|}{} & \textbf{GBM(8 hour)} & 5.6013 & 4.3693 & 60.50 & 71.99\\
\hline
\end{tabular}
\end{table*}

\section{ED Experience}
	A key differentiator of the work that we present here is that our prediction models were fully operationalized  into the clinical workflow, that of the ED charge nurse. Through collaborative design and planning sessions with ED nurses and other health system stakeholders, we developed an ED dashboard to surface the results of our predictions. Prediction based tools are often beset by difficulties in end-user understanding of probability based results \cite{jeffery2017participatory}.  Part of the solution to this problem is the early incorporation of end-user feedback and open discussions around tool utility.\\ Our dashboard was deployed for 6 months as part of pilot in a large suburban ED.  As part of this pilot, data quality was monitored continuously and multiple ML models were scored at 15 minute intervals.   
	End-user training was conducted during the pilot period. During this period charge nurses completed forms at the conclusion of each shift documenting their use of the dashboard and any actions the dashboard prompted (such as calling in additional staff for projected high load or sending staff home early for projected low load). In addition to the potential impact on nurse staffing, accurately forecasting ED arrivals and census may optimize care delivery in other ways - such as reducing waiting times, ED length of stay, and rates of patients leaving without being seen.  These additional key performance indicators (KPIs) were also be evaluated to determine the clinical utility of the deployed predictions. The iterative nature of this approach speaks to the engagement needs of the clinical end-users and the imperative of operationalizing machine learning in healthcare.  While accurate predictions are key to implementation success and end-user adoption, simple metrics such as prevalence of accuracy above a threshold (\textit{Accuracy $>$ 70\%}) will help health system stakeholders evaluate the impact and maintenance cost over a period of time.

\section{Discussion}
Our work demonstrates that subtle patterns in exogenous and endogenous variability in patient flow can be utilized to predict, with high accuracy, ED patient arrivals and census. 
Deployment of ML-based predictive models into a complex clinical workflow is challenging.  However, predicting ED census is an ideal ML healthcare problem to study for several reasons. First, predicting ED census every 15 minutes across 12 different models allows for $1,152$ predictions daily. Each prediction is clearly falsifiable with a measurable  outcome (the actual number of arrivals and patient census), and the follow-up interval is short (e.g., one must only wait 8 hours to determine the accuracy of all predictions). Second, many healthcare ML models are degraded by data censoring; for example, when predicting 30-day hospital readmissions, patients may avoid readmission, they may be readmitted at another facility. Additionally, according to the work of Jeffery and colleagues, prediction based tools are most useful when prompt decision and action are warranted by the end-users \cite{jeffery2017participatory}, however in some cases, such as predicting hospital readmissions, the action of the clinician can alter the outcome, thus making the prediction appear erroneous. In predicting ED load,  there are no actions that the users can take (other than the ED going on diversion status, which is done only seldom) that will alter the number of arrivals or census.  The large number of predictions, the short follow-up interval, and the availability of 'perfect information' about outcomes (akin to 'perfect information' games like chess) makes ED load prediction an ideal place to optimize model management processes.

We are continuing to improve the performance and clinical utility of these models by integrating additional data sources into our predictions. These sources can include events or include: local weather data, local sporting events, local traffic, local emergency medical services (EMS) activity, and Google Trends searches. We plan to further improve this solution by providing interpretability for the predictions to help ED staff make informed decisions. \cite{ahmad2018interpretable}

\bibliographystyle{aaai}
\bibliography{biblio}

\end{document}